\documentclass[11pt]{article}

\usepackage[margin=1in]{geometry}
\usepackage{booktabs}
\usepackage{microtype}
\usepackage{array}
\usepackage{xspace}
\usepackage{amsmath}
\usepackage{amssymb}
\usepackage{graphicx}
\usepackage{ragged2e}
\usepackage[round]{natbib}
\let\cite\citep
\usepackage[hidelinks]{hyperref}

\title{Coverage-Controlled Preference Mining from Noisy Claim Verification for Evidence-Grounded Generation}

\author{
Weixin Liu$^{1}$,
Congning Ni$^{2}$,
Qingyuan Song$^{1}$,
Susannah L. Rose$^{2}$,\\
Murat Kantarcioglu$^{3}$,
Bradley A. Malin$^{1,2}$,
Zhijun Yin$^{1,2}$\\[0.6em]
\small $^{1}$Vanderbilt University, Nashville, TN, USA\\
\small $^{2}$Vanderbilt University Medical Center, Nashville, TN, USA\\
\small $^{3}$Virginia Tech, Blacksburg, VA, USA
}

\date{}

\newcommand{\modelname}{\textsc{VERI-DPO}\xspace}
\newcommand{\dataset}{\textsc{VeriFact-BHC}\xspace}
\newcommand{\datasetiii}{\textsc{MIMIC-III-Ext-VeriFact-BHC}\xspace}
\newcommand{\datasetiv}{\textsc{MIMIC-IV-Ext-BHC}\xspace}

\begin{document}

\maketitle

\begin{abstract}
Evidence-grounded generation produces text whose claims should be supported by supplied evidence. Claim-level verifiers check generated claims against evidence and label each claim as Supported, Not Supported, or Not Addressed, making them scalable but imperfect feedback for factuality. Current uses often evaluate or rerank outputs; using verifier scores for training is harder because labels are noisy and apparent factuality can improve when a model produces shorter, vaguer summaries with fewer checkable claims. We study this problem in clinical Brief Hospital Course summarization, where outputs should be grounded in patient-specific EHR evidence. We introduce \modelname, a preference-mining framework that converts noisy claim verification into coverage-controlled summary-level preferences. For each evidence-window prompt, \modelname generates multiple candidate summaries and forms a preference pair from two summaries: a chosen summary with better aggregate verifier-estimated support and a rejected summary with worse support, while requiring comparable coverage, measured by verifiable summary content. Standard Direct Preference Optimization distills these pairs into a single-sample policy, avoiding inference-time reranking. On patient-disjoint \datasetiii test data, \modelname reduces Not Supported rates from 10.7\% to 1.9\% under the verifier used for preference mining and from 11.6\% to 6.4\% under a separately prompted GPT-4o judge. In 100 blinded pairwise assessments by two domain researchers, \modelname is preferred over the base model 56 times versus 18 times for factual faithfulness, with the remaining assessments being ties or other responses. In a locked zero-shot \datasetiv transfer test, where the 1,000-patient cohort, prompts, seeds, model artifacts, and analysis plan are fixed before evaluation and no model adaptation is allowed, \modelname lowers Not Supported rates with nearly unchanged scored-claim counts. Multi-seed ablations show that verifier-guided pair construction drives the gains, while coverage and anti-degeneration controls prevent apparent factuality improvements from coming from shorter or less checkable outputs.
\end{abstract}


\section{Introduction}
Summarization compresses long inputs into concise outputs that users can read, verify, and act on. In many settings, a useful summary must also be evidence-grounded: its factual claims should be supported by the supplied source evidence rather than merely be plausible. We study this problem in clinical Brief Hospital Course (BHC) summarization. A BHC is the discharge-summary section that describes what happened during a patient's hospital stay and supports handoff across care settings. This domain is a useful testbed because BHC summaries must be concise, clinically useful, and traceable to patient-specific electronic health record (EHR) evidence; unsupported statements may misstate diagnoses, treatments, procedures, or events relevant to follow-up care~\cite{hauschildt2022hospital,kripalani2007deficits}.

Large language models (LLMs) can generate readable clinical summaries, but their outputs may include claims that are unsupported, contradicted, or temporally misattributed~\cite{maynez2020faithfulness,kryscinski2020evaluating}. At the same time, avoiding unsupported claims by producing very short or vague summaries is not useful. We use \emph{coverage} to refer to the amount of checkable information retained in a summary, operationalized through scored claims and supported-claim counts. The objective is therefore not to optimize a binary notion of factuality, but to reduce unsupported content while preserving enough verifiable clinical information.

Claim-level verifiers provide a scalable way to evaluate this objective. A claim-level verifier checks each generated claim against evidence and predicts whether the claim is Supported, Not Supported, or Not Addressed. Such verifiers can be used for factuality evaluation~\cite{maynez2020faithfulness,kryscinski2020evaluating,chung2025mimic} and, when combined with retrieval or revision procedures, for inference-time correction~\cite{gao2023rarr,dhuliawala2024chain}. However, using verifier outputs as training supervision is more difficult. Individual claim labels can be noisy and retrieval-dependent, and naively preferring summaries with fewer detected errors can reward models for producing fewer checkable claims rather than more faithful summaries.

We therefore frame the central problem as preference construction from noisy evidence-linked feedback. In this paper, a preference pair consists of two candidate summaries generated for the same evidence-window prompt: a chosen summary $y^+$ and a rejected summary $y^-$. A useful pair should have a large aggregate verifier-estimated support difference between $y^+$ and $y^-$, while the two summaries should contain comparable amounts of checkable information. This summary-level construction avoids treating each verifier label as a ground-truth clinical judgment and instead uses the verifier to identify high-contrast comparisons that are less likely to be explained by omission.

\begin{figure}[t]
  \centering
  \includegraphics[width=\linewidth]{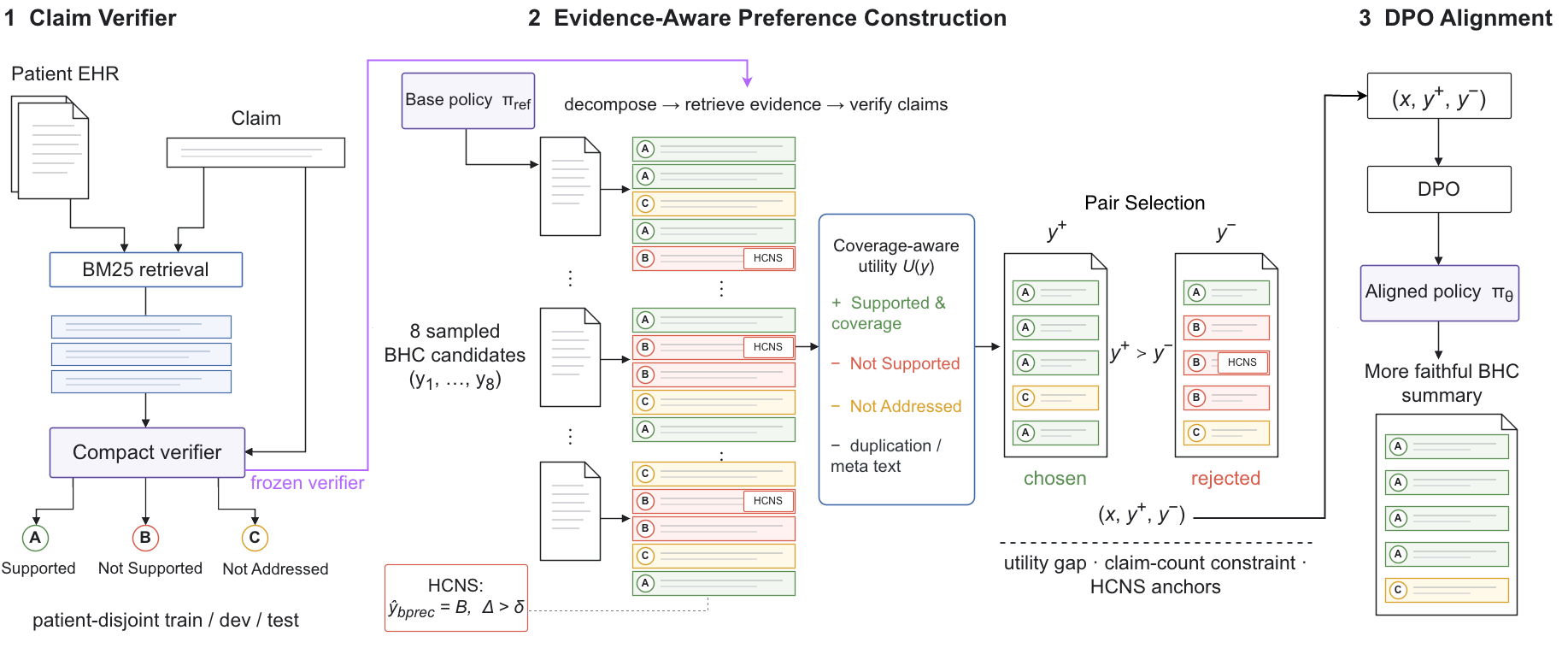}
  \caption{\textbf{VERI-DPO overview.}
  A compact verifier labels claims against patient-specific evidence as Supported, Not Supported, or Not Addressed. Candidate summaries are decomposed into claims, verified, scored by a coverage-aware utility, and filtered under utility-gap, claim-count, duplication, meta-text, and high-margin candidate Not Supported (HCNS) anchoring constraints to form $(x,y^+,y^-)$. Direct Preference Optimization (DPO) then distills these preferences into a single-sample policy. HCNS is a margin-qualified verifier signal used only for pair construction, not a clinically validated error label.}
  \label{fig:pipeline}
\end{figure}

To address this problem, we introduce \modelname, shown in Figure~\ref{fig:pipeline}. \modelname generates multiple candidate summaries for each evidence-window prompt, decomposes each candidate into claims, verifies those claims against patient-specific evidence, and scores each candidate with a coverage-aware utility. It then retains a pair only when the chosen summary has better aggregate verifier-estimated support than the rejected summary, while claim-count, repetition, and meta-text filters prevent degenerate wins from shorter or templated outputs. Standard Direct Preference Optimization (DPO)~\cite{rafailov2023direct} then distills these mined summary-level preferences into a single-sample summarizer, avoiding inference-time generation and reranking of multiple candidates.

We evaluate \modelname on patient-disjoint \datasetiii~\cite{chung2025mimic,johnson2016mimic} and a locked zero-shot \datasetiv transfer cohort~\cite{aali2025mimicivextbhc,aali2025datasetbenchmark}. The locked transfer protocol fixes the cohort, prompts, model artifacts, random seeds, verifier settings, and analysis plan before evaluation, with no MIMIC-IV-specific adaptation. Because automated judges are not clinical ground truth, we use them as directional factuality checks and complement them with blinded pairwise assessments by domain experts. Across automated judging, blinded domain-expert assessment, transfer evaluation, and matched ablations, the results show that verifier-guided pair construction improves factuality more reliably than generic preference optimization alone.

This paper makes three primary contributions. First, as a problem contribution, we formulate preference construction from noisy claim verification as a weak-supervision problem in which verifier-estimated factual contrast and information coverage must be controlled jointly. Second, as a method contribution, we introduce a coverage-controlled pair-mining framework that selects candidate-summary pairs with strong aggregate verifier-estimated support differences while filtering omission, repetition, meta-text, and claim-count degeneration. Third, as empirical evidence, we show in clinical long-form summarization that the resulting preferences improve factuality under automated judges, blinded domain-expert assessment, locked zero-shot transfer, and matched component ablations.

\section{Related Work}
This work connects three lines of research: evidence-grounded summarization and factuality evaluation, claim verification and inference-time correction, and preference alignment from automated feedback.

\paragraph{Evidence-grounded summarization and factuality.}
Instruction tuning~\cite{ouyang2022training} and retrieval-augmented generation~\cite{lewis2020retrieval} have improved generation systems, including long-form clinical summarization benchmarks~\cite{aali2025datasetbenchmark}, but factual reliability remains an obstacle. Standard reference-based metrics based on word or n-gram overlap can miss unsupported clinical claims, because a summary may be lexically similar to a reference while still adding information that is not supported by the source evidence~\cite{maynez2020faithfulness,kryscinski2020evaluating}. Proposition-level resources such as \dataset address this issue by decomposing BHC summaries into claims and labeling each claim against patient-specific EHR evidence~\cite{chung2025mimic}. We build on this line of work by using claim-evidence labels not only to evaluate summaries, but also to construct coverage-preserving training preferences.

\paragraph{Claim verification and inference-time correction.}
Fact verification commonly predicts whether a claim is supported, refuted, or not established by evidence~\cite{thorne2018fever,wadden2020fact}. Retrieval-and-revision methods such as RARR and Chain-of-Verification use retrieval, checking, or rewriting at inference time to reduce unsupported generation~\cite{gao2023rarr,dhuliawala2024chain}. In contrast, \modelname uses verifier outputs offline to construct preference pairs and trains a policy that produces one summary per prompt. We retain verifier-guided Best-of-8 reranking as an inference-intensive baseline that tests the value of verification without distilling it into the generator.

\paragraph{Preference alignment from automated feedback.}
Human-feedback methods and DPO optimize models from pairwise preferences~\cite{stiennon2020learning,rafailov2023direct}, and AI-generated feedback can reduce the cost of collecting such preferences~\cite{lee2023rlaif}. However, when automated preferences are based mainly on fewer detected errors, a model can be rewarded for producing shorter or vaguer outputs with fewer checkable claims. \modelname focuses on constructing evidence-based preference pairs with coverage and anti-degeneration controls, while using standard DPO rather than proposing a new preference-optimization objective.

\section{Method}
\subsection{Problem setup}
Consider an evidence-grounded generation task in which an output can be decomposed into verifiable claims. Given an evidence window $x$, the subset of source evidence supplied to the model for one generation prompt, instantiated here as patient-specific EHR evidence, a base summarizer samples candidate summaries $\{y_k\}_{k=1}^{K}$. Each candidate $y_k$ is decomposed into claims $\{c_{kj}\}_{j=1}^{m_k}$. A verifier $v_\phi$ receives a claim and retrieved evidence and predicts one of three labels: Supported ($A$), Not Supported ($B$), or Not Addressed ($C$). Given this information, our goal is to construct training examples $(x,y^+,y^-)$ whose preferred-response pair consists of two candidate summaries for the same evidence window: a chosen summary $y^+$ and a rejected summary $y^-$. The chosen summary should be more evidence-consistent than the rejected summary while retaining comparable information coverage. The key design choice is to use the verifier at the summary-pair level: individual claim predictions may be noisy, but aggregate, high-contrast comparisons can still provide useful alignment supervision. In simple terms, for each evidence-window prompt, we generate several candidate summaries, use verifier-derived aggregate scores and filtering rules to select a comparatively better and worse summary, and train on that pair only when the factuality difference is not explained by substantially lower information coverage.

\paragraph{Design desiderata.}
This problem imposes three requirements on preference construction. First, the method should be noise-tolerant. Individual verifier labels should guide preference mining, but should not be treated as ground-truth clinical labels. Second, the method should preserve coverage: a candidate should not be preferred merely because it is shorter, vaguer, or makes fewer checkable claims. Third, verifier guidance should be incorporated into the trained policy, rather than requiring inference-time Best-of-$K$ sampling and reranking. \modelname addresses these requirements by selecting aggregate high-contrast pairs, enforcing claim-count and coverage controls, and distilling the resulting preferences with standard DPO into a single-sample summarizer.

\subsection{Data and operational claim verifier}
To empirically evaluate our method, we use \datasetiii~\cite{chung2025mimic,johnson2016mimic}, a de-identified derivative of MIMIC-III containing 100 patients, 125 admissions, 4,787 source notes, and paired human- and LLM-written BHCs. By dataset construction, the source evidence available for verification excludes the final discharge summary, which contains the target BHC, reducing direct target leakage. The released dataset provides clinician annotations: BHCs are decomposed into 13,070 sentence- and atomic-level propositions and assigned majority-vote labels with adjudication~\cite{chung2025mimic}. Under this annotation scheme, Supported means that evidence establishes a claim; Not Supported indicates explicit conflict or contradiction; and Not Addressed indicates absent or insufficient evidence without a demonstrated contradiction. Because the distinction between Not Supported and Not Addressed is clinically important, Table~\ref{tab:label_examples} provides a small paraphrased example of how EHR evidence is translated into claim-level labels without reproducing patient text.

\begin{table}[t]
\caption{Paraphrased examples of claim-evidence labels. These examples are illustrative only and do not reproduce individual patient records.}
\label{tab:label_examples}
\centering
\scriptsize
\setlength{\tabcolsep}{3.0pt}
\begin{tabular}{@{}p{0.28\linewidth}p{0.17\linewidth}p{0.52\linewidth}@{}}
\toprule
\textbf{Claim} & \textbf{Label} & \textbf{Evidence interpretation} \\
\midrule
Started broad-spectrum antibiotics for pneumonia. & Supported & Medication and progress-note evidence documents antibiotic treatment for pneumonia. \\
Creatinine improved after fluids. & Not Addressed & The evidence includes fluids and creatinine values, but does not establish the claimed improving trend. \\
Underwent coronary bypass surgery during admission. & Not Supported & Procedure and cardiology evidence explicitly contradicts the stated surgery or documents an alternative procedure. \\
\bottomrule
\end{tabular}
\end{table}

We split the patient records into 72/8/20 train/development/test sets, yielding 9,217/895/2,644 verifier instances after claim-validity filtering. We excluded 314 propositions because they could not be deterministically converted into standalone claim-evidence verification instances. Retrieval is patient-specific and never crosses splits. Hyperparameters, checkpoints, and operating points are selected using training/development patients only, while test patients are reserved for final evaluation. To reduce target leakage, discharge-summary source row identifiers are removed before indexing each patient's evidence corpus. Two-stage Okapi BM25 retrieval~\cite{robertson2009probabilistic}, a term-matching retrieval function with document-length normalization, ranks notes and then sentence-like evidence units, with near-duplicate removal and a cap per source note. 

The operational verifier is Med42-8B~\cite{christophe2024med42}, fine-tuned with Low-Rank Adaptation (LoRA)~\cite{hu2022lora} and 4-bit NormalFloat4 (NF4) quantization~\cite{dettmers2023qlora}. A fixed prompt contains the label policy, retrieved evidence, and the claim, using the single-token labels \texttt{``A''}, \texttt{``B''}, and \texttt{``C''} for Supported, Not Supported, and Not Addressed, respectively; the loss is applied only to these leading-space label tokens. The operational checkpoint and Not Supported logit bias are selected from development-set verifier and mining diagnostics only; test summaries and GPT-4o results are not used for selection. 

A high-margin candidate Not Supported signal (HCNS) is defined as
\begin{equation}
\mathrm{HCNS}\iff \hat y_{b_{\mathrm{prec}}}=B
\ \wedge\
\Delta>\delta,
\qquad
\Delta=\ell_B-\max(\ell_A,\ell_C),
\label{eq:hcns}
\end{equation}
where $(\ell_A,\ell_B,\ell_C)$ are verifier logits, $b_{\mathrm{prec}}=-0.34$ is a more selective development-selected bias, and $\delta=0.8$. Development auditing gives approximately 0.29 precision and 0.18 recall for individual HCNS predictions, compared with an approximately 8\% Not Supported base rate among development claim-verification instances. Thus, HCNS enriches for likely Not Supported content by about 3.6-fold, but most individual HCNS flags remain false positives. We therefore use HCNS only as a noisy anchor inside the pair-selection rule defined below, where it is combined with aggregate verifier separation, claim-count matching, chosen-side error caps, rejected-side anchoring, and anti-degeneration filters, rather than as a proposition-level clinical error label. The unit of supervision is a filtered summary-level contrast whose aggregate verifier separation and coverage properties can be audited before DPO training.

\subsection{Coverage-controlled preference construction}
For each of the 72 training patient records, we sample 30 evidence-window prompts and generate eight BHC candidates per prompt from the base policy using nucleus sampling~\cite{holtzman2020curious}. Each candidate is segmented into sentence-level claims; patient-specific evidence is retrieved for each claim; and the operational verifier supplies label counts $(n_A,n_B,n_C)$. Let $n=n_A+n_B+n_C$. We score a candidate summary by
\begin{equation}
\begin{split}
U(y)=&\ \lambda_A n_A-\lambda_B n_B-\lambda_C n_C
+\lambda_{\mathrm{cov}}\min(n,n_0)\\
&-\lambda_{\mathrm{dup}}(\mathrm{dup\_frac}\cdot n)
-\lambda_{\mathrm{meta}}\,\mathrm{meta\_hits}.
\end{split}
\label{eq:utility}
\end{equation}
The factual terms reward supported claims and penalize Not Supported claims most strongly; the Not Addressed penalty discourages unsupported-but-not-contradicted content. The coverage reward discourages trivially short summaries, while the duplication and meta-text penalties discourage repetitive or templated degeneration. We treat the coefficients as preference-mining hyperparameters that encode this priority ordering: Not Supported errors receive the largest penalty, coverage is rewarded up to a cap, and repetition or meta-text is penalized. In this paper, we use $(\lambda_A,\lambda_B,\lambda_C,\lambda_{\mathrm{cov}},\lambda_{\mathrm{dup}},\lambda_{\mathrm{meta}})=(1,3,0.5,0.25,2,2)$ and $n_0=12$, selected from development mining diagnostics.

A prompt yields a retained pair $(y^+,y^-)$ only if all of the following hold: (i) $U(y^+)-U(y^-)\ge2$; (ii) the two summaries differ by at most six scored claims; (iii) $y^+$ contains at most one HCNS and at most two predicted-$B$ claims; and (iv) $y^-$ contains at least one HCNS. The claim-count constraint prevents choosing a summary merely because it says less, and the anchoring constraints require the rejected candidate to contain at least one margin-qualified contradiction signal. This procedure yields 1,513 training pairs. In 98.02\% of pairs, the chosen candidate has fewer predicted-$B$ claims than the rejected candidate; their mean predicted-$B$ counts are 0.484 and 4.767, respectively, corresponding to a rejected-minus-chosen gap of 4.283. Table~\ref{tab:pref_mining_diagnostics} summarizes the retained training pairs, and Table~\ref{alg:mining} provides an algorithm-style summary of the mining procedure. These diagnostics are not used as test evidence; they show that the mined data provide high-contrast summary-level preferences before DPO training.

\begin{table}[t]
\caption{Preference-mining diagnostics on the training split. The table reports only training-pair properties, not held-out evaluation metrics.}
\label{tab:pref_mining_diagnostics}
\centering
\scriptsize
\setlength{\tabcolsep}{4.0pt}
\begin{tabular}{@{}lc@{}}
\toprule
\textbf{Diagnostic} & \textbf{Value} \\
\midrule
Training patients / prompts per patient / candidates per prompt & 72 / 30 / 8 \\
Retained preference pairs & 1,513 \\
$b_{\mathrm{prec}}$ / HCNS margin $\delta$ & $-0.34$ / 0.8 \\
Chosen has fewer predicted-$B$ claims than rejected & 98.02\% \\
Mean predicted-$B$ claims, chosen / rejected & 0.484 / 4.767 \\
Rejected-minus-chosen predicted-$B$ gap & 4.283 \\
Mean utility gap $U(y^+)-U(y^-)$ & 20.30 \\
\bottomrule
\end{tabular}
\end{table}

\begin{table}[t]
\caption{Algorithm-style summary of evidence-aware preference mining in \modelname.}
\label{alg:mining}
\centering
\scriptsize
\setlength{\tabcolsep}{2.5pt}
\begin{tabular}{@{}p{0.045\linewidth}p{0.90\linewidth}@{}}
\toprule
\multicolumn{2}{@{}p{0.945\linewidth}@{}}{\textbf{Input:} training patients $\mathcal{P}_{\mathrm{train}}$, base policy $\pi_0$, verifier $v_\phi$, candidates per prompt $K$, utility $U$, thresholds $(\tau_U,\tau_n,\tau_B,\tau_{\mathrm{HCNS}})$.}\\
\multicolumn{2}{@{}p{0.945\linewidth}@{}}{\textbf{Output:} preference dataset $\mathcal{D}=\{(x,y^+,y^-)\}$.}\\
\midrule
1 & Initialize $\mathcal{D}\gets\emptyset$.\\
2 & For each training patient and sampled evidence-window prompt $x$, sample $K$ candidate summaries $\{y_k\}_{k=1}^{K}\sim\pi_0(\cdot\mid x)$.\\
3 & For each candidate $y_k$, decompose it into claims and retrieve patient-specific evidence for each claim.\\
4 & Apply $v_\phi$ to obtain $(n_A,n_B,n_C)$, HCNS count, duplication, and meta-text counts; compute $U(y_k)$ using Eq.~\eqref{eq:utility}.\\
5 & Select $(y^+,y^-)$ only if $U(y^+)-U(y^-)\ge\tau_U$ and $|n(y^+)-n(y^-)|\le\tau_n$.\\
6 & Require $y^+$ to satisfy chosen-side $B$/HCNS caps and $y^-$ to contain at least one HCNS.\\
7 & If all constraints hold, add $(x,y^+,y^-)$ to $\mathcal{D}$; otherwise discard the prompt.\\
\bottomrule
\end{tabular}
\end{table}

\subsection{DPO alignment and baselines}
Let $\pi_\theta$ be the trainable policy and $\pi_{\mathrm{ref}}$ the frozen base policy. We optimize the standard DPO objective~\cite{rafailov2023direct}
\begin{equation}
\mathcal L_{\mathrm{DPO}}=-\mathbb E_{(x,y^+,y^-)}
\log\sigma\!\left(\beta\left[
\log\frac{\pi_\theta(y^+\mid x)}{\pi_{\mathrm{ref}}(y^+\mid x)}-
\log\frac{\pi_\theta(y^-\mid x)}{\pi_{\mathrm{ref}}(y^-\mid x)}
\right]\right).
\label{eq:dpo}
\end{equation}

The summarizer is Llama-3.1-8B-Instruct~\cite{grattafiori2024llama}, trained with LoRA ($r=16$, $\alpha=32$) with a learning rate of $10^{-5}$. We evaluate DPO checkpoints only on the 48 development prompts; GPT-4o and held-out test data are not used for selection.

Checkpoint selection minimizes local-verifier Not Supported rate (NS-rate) subject to three prespecified anti-degeneration gates: at least 80\% output-quality pass rate, mean character length at least 95\% of Base, and mean scored-claim count no more than 0.5 claims lower than Base. This procedure selects $\beta=0.05$ and three epochs. On the development subset, the selected checkpoint obtains a local NS-rate of 1.63\%, compared with 10.41\% for Base and 11.81\% for SFT, while satisfying all three gates.

We compare \modelname against three baselines. First, Base is the unaligned summarizer. Second, SFT trains on the chosen responses only and controls for learning the chosen-output distribution without pairwise preference pressure. Third, Verifier-guided Best-of-8 reranking samples eight base candidates at inference time and selects the candidate maximizing $R(y)=n_A-1.10n_B$. Best-of-8 is a strong inference-intensive baseline because it requires eight generations and verifier passes for each prompt, whereas \modelname produces one summary.

\section{Experimental Design}
The primary inferential factuality endpoint is patient-level macro NS-rate. For each patient, we compute prompt-level NS-rate as the fraction of scored claims labeled Not Supported and then average these prompt-level rates across that patient's prompts. We report descriptive complete-set rates over the full aligned prompt set, while confidence intervals and hypothesis tests use patient-level paired differences. Mean Not Supported and Supported claims, scored claims, output-quality pass rate, duplication, and output length are reported as anti-degeneration diagnostics. Because prompts are nested within patients, inferential comparisons use the patient rather than the prompt or claim as the paired unit.

\subsection{Primary held-out evaluation}
All methods are evaluated on the same 120 prompts from 20 held-out \datasetiii patients, with patient identifiers, prompt keys, generation identifiers, and prompt hashes aligned across systems. The local operational verifier performs per-claim retrieval from the patient corpus. GPT-4o~\cite{hurst2024gpt} is used only as a second automated cross-judge: it judges each claim against the labeled evidence excerpts supplied to the corresponding generation prompt, with model identity and local-verifier outputs hidden. GPT-4o is not used for pair construction, operating-point selection, checkpoint selection, or training. Because the local verifier and GPT-4o use different evidence-construction procedures, we use GPT-4o as an automated sensitivity check: within each judge, we compare Base and \modelname on the same prompts, but we do not compare absolute NS-rates across the two judges as if they were calibrated to the same scale.

Claim metrics and output-length diagnostics are computed on the same 120-prompt set for every method. For paired inference, we average prompt-level outcomes within each patient and compute nonparametric patient-level bootstrap confidence intervals, sign-flip permutation tests, and Wilcoxon signed-rank tests.

\subsection{Blinded pairwise assessment by domain researchers}
Automated judges cannot establish holistic report quality, so we additionally conduct a blinded pairwise assessment by two domain researchers with expertise in clinical text generation and medical report summarization. They independently evaluate 50 Base-\modelname pairs spanning all 20 test patients. Ten patients contribute three pairs and ten contribute two; cases are sampled without using automated-judge scores. Both summaries are presented with the same evidence, while system identities, automated scores, and reference BHCs are hidden. A/B order is balanced within evaluator and reversed between evaluators.

The prespecified primary endpoint is factual faithfulness; secondary endpoints are medically important information preservation and overall medical-report draft preference. Directional choices are coded $+1$ for \modelname and $-1$ for Base; equivalent/no-preference categories are coded zero, while insufficient-evidence responses are excluded from the corresponding directional mean. Scores are averaged across evaluators and sampled prompts within patient. We report response counts, directional win rates, patient-level bootstrap intervals, sign-flip and Wilcoxon tests, exact agreement, and unweighted Cohen's $\kappa$~\cite{cohen1960coefficient}; Holm correction~\cite{holm1979simple} is applied across the three endpoint-specific sign-flip tests.

\subsection{Locked zero-shot MIMIC-IV transfer}
To test whether the paired direction persists beyond the small held-out \datasetiii set, we perform zero-shot transfer to \datasetiv v1.2.0~\cite{aali2025mimicivextbhc,aali2025datasetbenchmark}, derived from MIMIC-IV-Note~\cite{johnson2023mimicivnote} and linked to MIMIC-IV metadata~\cite{johnson2023mimiciv}. Because dates are patient-shifted, retained admissions are conservatively inferred to occur after 2012, making the transfer cohort temporally later than the MIMIC-III-derived training and evaluation data without relying on exact calendar dates. Prespecified filters exclude residual BHC headings, exact target containment, and exact cross-patient input-target duplicates, and retain one admission per patient. We use ``locked'' to mean that the evaluation cohort and all analysis decisions are finalized before inspecting any system results on the transfer cohort. The locked cohort contains 1,000 unique patients.

Before evaluation, we finalize cohort selection, target-blind evidence extraction, model artifacts, paired random seeds, generation controls, verifier settings, claim tasks, and analysis scripts. Base and \modelname receive identical prompts and evidence. No MIMIC-IV-specific training, prompt tuning, or post-opening scientific modification is performed. The finalized splitter produces 10,042 Base and 10,029 \modelname claims. Both judges evaluate the same fixed collection of 20,071 system-generated claim instances; exact duplicate claim-evidence tasks are judged once and mapped back, yielding 19,860 unique GPT-4o requests.

\subsection{Matched component ablations}
We compare four preference-construction arms. \textbf{Full} uses verifier-derived pair orientation, HCNS/$B$ anchoring, and all utility-level controls. \textbf{Random-pair DPO} randomly selects two valid candidates from the same pool and randomly assigns their direction; it is a negative control for generic DPO optimization. \textbf{No HCNS/$B$ anchoring} retains the full utility and claim-count matching but removes chosen-side HCNS/$B$ caps and the rejected-side HCNS requirement. \textbf{No utility/matching controls} retains verifier orientation and anchoring but removes coverage, repetition, meta-text, and claim-count-matching terms, ranking candidates by $(\lambda_A n_A-\lambda_B n_B-\lambda_C n_C)/\max(n,1)$ with a gap threshold of 0.15 rather than 2.0.

All arms use the same 1,513 prompt keys, training patients, candidate pools, backbone, DPO configuration, and test prompts. Each arm is trained with seeds 42, 43, and 44, yielding 12 policies. The ablation Full arm is a matched three-seed reconstruction used only for component analysis and is distinct from the development-selected main \modelname policy reported in Table~\ref{tab:main_results}. For each prespecified Full-versus-ablation contrast, claim outcomes are averaged within patient and then across seeds before paired inference.

\section{Results}
\subsection{Main factuality and anti-degeneration results}
Table~\ref{tab:main_results} reports descriptive system-level results on the complete 120-prompt held-out set. The NS-rates in this table are computed over scored claims on the full aligned prompt set; patient-level macro paired inference is reported below. Under the mining-aligned local verifier, \modelname reduces descriptive NS-rate from 10.7\% to 1.9\% and mean Not Supported claims from 1.98 to 0.36. Under GPT-4o, the corresponding reductions are 11.6\% to 6.4\% and 2.14 to 1.26. Supported-claim counts increase under both judges, and outputs are longer. SFT provides little improvement, suggesting that the gains are not explained by learning the chosen-response format alone. Best-of-8 is competitive but inference-intensive, while \modelname is a single-sample policy.

\begin{table}[t]
\caption{Main results on the complete held-out set of 120 aligned prompts from 20 held-out patients. NS-rates in the table are descriptive complete-set rates computed over scored claims on the identical aligned prompt set; patient-level macro paired inference is reported in text.}
\label{tab:main_results}
\centering
\scriptsize
\setlength{\tabcolsep}{3.6pt}
\begin{tabular}{@{}lccccccc@{}}
\toprule
& \multicolumn{3}{c}{\textbf{Local verifier}} &
\multicolumn{3}{c}{\textbf{GPT-4o}} &
\multicolumn{1}{c}{\textbf{Overall}} \\
\cmidrule(lr){2-4}\cmidrule(lr){5-7}\cmidrule(lr){8-8}
\textbf{Method} & \textbf{NS-rate$\downarrow$} & \textbf{\#NS$\downarrow$} & \textbf{\#Supp.$\uparrow$} &
\textbf{NS-rate$\downarrow$} & \textbf{\#NS$\downarrow$} & \textbf{\#Supp.$\uparrow$} &
\textbf{Chars$\uparrow$} \\
\midrule
Base & 10.7\% & 1.98 & 15.0 & 11.6\% & 2.14 & 12.0 & 1855 \\
SFT & 10.1\% & 1.92 & 14.4 & 10.0\% & 2.04 & 12.8 & 1865 \\
Best-of-8 & 3.4\% & 0.69 & 18.9 & 8.3\% & 1.76 & 14.4 & 1900 \\
\textbf{VERI-DPO} & \textbf{1.9\%} & \textbf{0.36} & \textbf{19.1} & \textbf{6.4\%} & \textbf{1.26} & \textbf{14.6} & \textbf{2159} \\
\bottomrule
\end{tabular}
\end{table}

For Base-\modelname paired inference, all 20 held-out patients contribute to the patient-level comparison. The local patient-level NS-rate difference is $-8.68$ percentage points (95\% confidence interval [CI] $[-11.08,-6.23]$, sign-flip $p=3.0\times10^{-5}$), and the GPT-4o difference is $-5.78$ points (95\% CI $[-8.03,-3.40]$, $p=2.9\times10^{-4}$). Eighteen of 20 patients show lower NS-rates under each judge. Local Supported claims increase by 4.13 per summary, and mean length increases by 304 characters.

\subsection{Cross-judge and human assessment}
The GPT-4o results in Table~\ref{tab:main_results} test whether improvements persist beyond the mining-aligned verifier. To further test whether lower automated NS-rates correspond to holistic report quality, we conduct a blinded pairwise assessment by domain researchers. Across 100 evaluator-pair assessments, \modelname is preferred over Base 56 to 18 for factual faithfulness, 47 to 20 for information preservation, and 51 to 19 for overall draft preference (Table~\ref{tab:human}). All three patient-level effects remain significant after Holm correction. Exact agreement is 58-64\% ($\kappa=0.358$-$0.406$), and both evaluators independently favor \modelname on every endpoint.

\begin{table}[t]
\caption{Blinded pairwise assessment of 50 Base-\modelname outputs by two domain researchers. Counts are over 100 assessments; directional win rates exclude non-directional responses. Patient net-preference scores range from $-1$ (Base favored) to $+1$ (\modelname favored).}
\label{tab:human}
\centering
\scriptsize
\setlength{\tabcolsep}{2.0pt}
\renewcommand{\arraystretch}{0.95}
\begin{tabular*}{0.96\linewidth}{@{\extracolsep{\fill}}p{0.24\linewidth}cccccc@{}}
\toprule
\textbf{Endpoint} 
& \textbf{VERI-DPO} 
& \textbf{Base} 
& \shortstack{\textbf{Tie/}\\\textbf{other}} 
& \shortstack{\textbf{VERI-DPO}\\\textbf{win}$\uparrow$} 
& \shortstack{\textbf{Patient}\\\textbf{mean}$\uparrow$} 
& \shortstack{\textbf{95\% CI;}\\\textbf{adj. $p$}} \\
\midrule
Factual faithfulness & \textbf{56} & 18 & 26 & \textbf{75.7\%} & \textbf{$+0.388$} & $[+0.167,+0.588]$; 0.0117 \\
Information preservation & \textbf{47} & 20 & 33 & \textbf{70.1\%} & \textbf{$+0.286$} & $[+0.098,+0.457]$; 0.0182 \\
Overall draft preference & \textbf{51} & 19 & 30 & \textbf{72.9\%} & \textbf{$+0.313$} & $[+0.108,+0.525]$; 0.0182 \\
\bottomrule
\end{tabular*}
\end{table}

\subsection{Locked zero-shot MIMIC-IV transfer}
The policies selected before transfer evaluation are evaluated on all 1,000 \datasetiv patients. At the prespecified local operating point, macro NS-rate falls from 1.94\% to 0.93\%; GPT-4o shows a decrease from 11.17\% to 10.22\% (Table~\ref{tab:mimiciv}). Scored claims per summary are essentially unchanged ($-0.013$, 95\% CI $[-0.033,+0.007]$), although controlled outputs are modestly shorter. The improvement is smaller than on \datasetiii, but its value is protocol independence: the cohort is opened once, the policy is not adapted, and paired systems share prompts, evidence, seeds, and analysis procedures.

\begin{table}[t]
\caption{Locked zero-shot evaluation on 1,000 conservatively inferred post-2012 \datasetiv patients. Both judges evaluate the identical fixed set of 20,071 system-generated claim instances; differences are \modelname minus Base.}
\label{tab:mimiciv}
\centering
\scriptsize
\setlength{\tabcolsep}{4.2pt}
\begin{tabular}{@{}lcccc@{}}
\toprule
\textbf{Metric} & \textbf{Base} & \textbf{VERI-DPO} & \textbf{Difference} & \textbf{Bootstrap 95\% CI} \\
\midrule
\multicolumn{5}{@{}l}{\textit{Local operational verifier}} \\
Not Supported rate$\downarrow$ & 1.94\% & \textbf{0.93\%} & $-1.00$ pp & $[-1.33,-0.69]$ \\
Supported rate$\uparrow$ & 93.64\% & \textbf{98.22\%} & $+4.58$ pp & $[+4.03,+5.14]$ \\
Not Addressed rate$\downarrow$ & 4.42\% & \textbf{0.85\%} & $-3.58$ pp & $[-4.04,-3.13]$ \\
\addlinespace
\multicolumn{5}{@{}l}{\textit{GPT-4o cross-judge}} \\
Not Supported rate$\downarrow$ & 11.17\% & \textbf{10.22\%} & $-0.95$ pp & $[-1.28,-0.63]$ \\
Supported rate$\uparrow$ & 81.67\% & \textbf{84.96\%} & $+3.30$ pp & $[+2.79,+3.79]$ \\
Not Addressed rate$\downarrow$ & 7.16\% & \textbf{4.82\%} & $-2.34$ pp & $[-2.70,-1.99]$ \\
\addlinespace
\multicolumn{5}{@{}l}{\textit{Judge-independent output controls}} \\
Scored claims / summary & 10.042 & 10.029 & $-0.013$ & $[-0.033,+0.007]$ \\
Controlled words & 153.29 & 147.12 & $-6.17$ & $[-7.87,-4.46]$ \\
Controlled characters & 1006.03 & 988.14 & $-17.90$ & $[-28.86,-7.14]$ \\
\bottomrule
\end{tabular}
\end{table}

At the summary level, the proportion with at least one Not Supported claim falls from 17.8\% to 9.0\% locally and from 58.5\% to 54.4\% under GPT-4o. The fraction with at least one GPT-4o Not Supported or Not Addressed claim decreases from 75.9\% to 67.6\% (paired difference $-8.3$ pp, 95\% CI $[-10.5,-6.1]$). Local paired NS-rate reductions remain negative at all prespecified operating points. Because MIMIC-III and MIMIC-IV belong to the MIMIC/Beth Israel lineage~\cite{johnson2016mimic,johnson2023mimiciv}, these results provide temporal and dataset-version evidence rather than cross-institutional validation.

\subsection{Component analysis}
Before training the ablated policies, the matched pair pools already show the intended mechanism changes (Table~\ref{tab:matched_pairs_main}). Full and No anchoring produce similar rejected-minus-chosen predicted-$B$ gaps, whereas Random-pair DPO largely removes verifier-guided orientation: only 2.64\% of Random pairs match the Full ordered pair and unordered overlap is 6.41\%. No utility/matching produces the largest verifier gap, but its chosen summaries have fewer claims than rejected summaries, illustrating why verifier separation alone is not enough.

\begin{table}[t]
\caption{Pair-level diagnostics for the matched component ablation before DPO training. Each arm contains the same 1,513 prompt keys. $\Delta B$ is rejected minus chosen; $\Delta$chars and $\Delta$claims are chosen minus rejected; overlap is computed relative to Full.}
\label{tab:matched_pairs_main}
\centering
\scriptsize
\setlength{\tabcolsep}{2.8pt}
\resizebox{\linewidth}{!}{%
\begin{tabular}{@{}lrrrrrrr@{}}
\toprule
\textbf{Arm} & \textbf{Chosen \#B} & \textbf{Rejected \#B} & $\boldsymbol{\Delta B}$ & \textbf{Chosen fewer B} & $\boldsymbol{\Delta}$\textbf{chars} & $\boldsymbol{\Delta}$\textbf{claims} & \textbf{Unordered overlap} \\
\midrule
Full VERI-DPO & 0.484 & 4.767 & 4.283 & 98.02\% & +17.38 & +0.126 & 100.00\% \\
Random-pair DPO & 2.199 & 2.275 & 0.076 & 39.26\% & -1.29 & -0.089 & 6.41\% \\
No HCNS/\texttt{B} anchoring & 0.679 & 4.979 & 4.300 & 93.39\% & +21.97 & +0.368 & 82.09\% \\
No utility/matching controls & 0.343 & 5.095 & 4.752 & 99.41\% & -7.90 & -1.557 & 40.38\% \\
\bottomrule
\end{tabular}}
\end{table}

Table~\ref{tab:ablation} reports the matched three-seed component ablation. Random-pair construction increases local NS-rate from $1.34\pm0.30$\% to $10.33\pm0.79$\% and GPT-4o NS-rate from $7.39\pm0.45$\% to $10.16\pm1.12$\%, showing that the gains are not a generic consequence of DPO optimization. Removing utility/matching increases repetition, reduces supported claims, and worsens GPT-4o factuality despite producing an even larger pair-level verifier gap before training. Thus, verifier separation alone is insufficient without anti-degeneration controls. Removing HCNS/$B$ anchoring has a smaller and judge-dependent effect: Full is more conservative under the mining verifier, but GPT-4o does not confirm an independent factuality advantage and assigns more Supported content without anchoring. We therefore interpret anchoring as a conservative, mining-aligned operating choice rather than a universally necessary component.

\begin{table}[t]
\caption{Matched three-seed preference-construction ablation. Values are mean $\pm$ standard deviation (SD) across independently trained policies. Table rates are descriptive complete-set averages on the same 120 held-out prompts; patient-level paired contrasts are reported in text. Random-pair DPO is a negative control for generic DPO training; No utility/matching tests whether verifier orientation without coverage controls is sufficient.}
\label{tab:ablation}
\centering
\scriptsize
\setlength{\tabcolsep}{3.4pt}
\begin{tabular}{@{}lccccc@{}}
\toprule
\textbf{Method} & \textbf{Local NS$\downarrow$} & \textbf{GPT NS$\downarrow$} & \textbf{GPT \#NS$\downarrow$} & \textbf{GPT \#Supp.$\uparrow$} & \textbf{Dup.$\downarrow$} \\
\midrule
Full & \textbf{1.34$\pm$0.30\%} & 7.39$\pm$0.45\% & \textbf{1.333$\pm$0.079} & 13.11$\pm$0.17 & 0.0178$\pm$0.0078 \\
Random pairs & 10.33$\pm$0.79\% & 10.16$\pm$1.12\% & 1.889$\pm$0.248 & 11.59$\pm$0.37 & 0.0840$\pm$0.0092 \\
No anchoring & 1.97$\pm$0.50\% & \textbf{7.05$\pm$0.28\%} & 1.359$\pm$0.043 & \textbf{13.89$\pm$0.68} & \textbf{0.0082$\pm$0.0013} \\
No utility/matching & 1.56$\pm$0.23\% & 8.56$\pm$0.67\% & 1.569$\pm$0.117 & 12.36$\pm$0.46 & 0.0904$\pm$0.0093 \\
\bottomrule
\end{tabular}
\end{table}

Patient-level inference confirms these patterns. Relative to Random, Full lowers NS-rate by $-8.65$ pp locally (95\% CI $[-10.49,-6.76]$) and $-3.29$ pp under GPT-4o (95\% CI $[-4.78,-1.92]$), with $+1.911$ GPT-4o Supported claims. Relative to No utility/matching, Full improves GPT-4o NS-rate by $-1.63$ pp (95\% CI $[-2.84,-0.52]$) and preserves more Supported content, while duplication is 0.0178 versus 0.0904. Relative to No anchoring, Full improves local NS-rate by $-0.74$ pp, but the GPT-4o NS-rate contrast is not significant. Full matched pairs have a rejected-minus-chosen predicted-$B$ gap of 4.283, whereas Random largely removes the construction signal ($\Delta B=0.076$; unordered overlap with Full pairs 6.41\%).

\section{Discussion}
The experiments suggest that the useful supervision does not come from DPO alone or from treating verifier labels as ground truth. Rather, it comes from constructing summary-level comparisons in which verifier-estimated factuality differs substantially while information coverage remains similar. Random pairing removes the factual direction of this supervision, whereas removing coverage and anti-degeneration controls preserves verifier separation but can encourage less informative or more repetitive outputs. These findings support preference construction, rather than the DPO objective itself, as the main mechanism: noisy verifier feedback becomes useful only after it is aggregated, filtered, and constrained into coverage-controlled preference pairs.

The largest improvement occurs under the verifier used to construct preferences, so judge coupling and circular evaluation cannot be ruled out for the local-verifier endpoint. We therefore treat the local verifier as a mining-aligned diagnostic rather than as independent clinical validation. However, the paired direction is also observed under the separately prompted GPT-4o cross-judge, blinded pairwise assessment by domain researchers, and locked zero-shot \datasetiv transfer. These checks do not make the automated labels interchangeable with expert ground truth, but they reduce the likelihood that the result is only a local-verifier artifact.

The results also argue against a pure omission explanation. On the primary held-out set, \modelname increases Supported claims and output length on the complete aligned prompt set. In the locked transfer, scored claim counts are essentially unchanged. The No utility/matching ablation further shows that a larger verifier gap alone can produce worse repetition and lower independently supported content, supporting the need for explicit coverage and anti-degeneration controls. Compared with verifier-guided Best-of-8 reranking, the aligned policy produces one summary per prompt rather than requiring multiple generations and verifier passes at inference time.

The zero-shot transfer gain is smaller than the primary held-out gain, but its value lies in protocol independence: the cohort is evaluated once, the policy is not adapted, systems share evidence and paired seeds, and the cross-judge evaluates the same fixed claim tasks. The useful unit of supervision appears to be a filtered summary-level contrast rather than an individually reliable verifier label. Establishing whether this observation generalizes beyond BHC summarization requires additional evidence-grounded generation tasks, model families, retrieval settings, and institutions.

\paragraph{Limitations.}
The primary held-out set contains only 20 patients, although inference is patient-level and transfer is tested on 1,000 additional patients. The human evaluators are domain researchers rather than practicing clinicians; larger blinded physician studies are needed before clinical-deployment claims. The \datasetiv evaluation uses automated judges only, and MIMIC-III to MIMIC-IV is temporal/dataset-version transfer rather than cross-institutional validation. Results may depend on retrieval quality, evidence-window scope, verifier calibration, and the definitions of Not Supported and Not Addressed. The main summarizer uses one backbone, and the no-anchoring ablation removes a bundle of constraints rather than isolating HCNS alone. GPT-4o remains an automated cross-judge, not external clinical ground truth~\cite{zheng2023judging}.

\paragraph{Reproducibility and audit trail.}
Patient splits are disjoint; pair mining uses training patients only; hyperparameter, checkpoint, and operating-point selection use development patients; and the test split is not accessed for preference construction. Main outputs are complete for the 120-prompt held-out set and are aligned by de-identified patient identifier, prompt key, generation identifier, and prompt hash. Matched ablations use identical prompt keys and SHA-1 prompt hashes across all 12 policies. Human-evaluation and locked-transfer manifests, mappings, model artifacts, evidence selections, paired seeds, claim tasks, and analysis specifications are finalized and hash-identified before evaluation. Prompt templates, configurations, hashed manifests, and analysis scripts are separable from patient-derived text and will be shared where permitted; patient-derived text remains governed by PhysioNet credentialing and data-use requirements~\cite{goldberger2000physiobank}.

\section{Conclusion}
We show that noisy claim verification can provide useful preference supervision when it is aggregated into high-contrast, coverage-matched summary pairs rather than treated as ground-truth labels. In clinical long-form summarization, these pairs enable standard DPO to reduce unsupported content without reducing the number of scored claims, and the direction of improvement persists under a separate automated judge, blinded assessment by domain researchers, and locked zero-shot MIMIC-IV transfer. The main contribution is therefore not a new DPO objective, but a method for constructing more reliable, coverage-controlled preference data from imperfect evidence-linked feedback.

\bibliographystyle{plainnat}
\bibliography{references}
\end{document}